\documentclass[10pt,twocolumn,letterpaper]{article}

\usepackage{wacv}              

\usepackage[table]{xcolor}
\usepackage{gensymb}
\usepackage{amssymb}
\usepackage{pifont}
\definecolor{wacvblue}{rgb}{0.21,0.49,0.74}
\usepackage[pagebackref,breaklinks,colorlinks,allcolors=wacvblue]{hyperref}

\def\wacvPaperID{2318} 
\def\confName{WACV}
\def\confYear{2026}

\title{KD360-VoxelBEV: LiDAR and 360-degree Camera Cross Modality Knowledge Distillation for Bird's-Eye-View Segmentation}

\author{
Wenke E\thanks{Equal contribution.} \quad
Yixin Sun\footnotemark[1] \quad
Jiaxu Liu \quad
Hubert P. H. Shum \\
Amir Atapour-Abarghouei \quad
Toby P. Breckon \\
Durham University, UK
}

\begin{document}
\maketitle
\begin{abstract}
\label{sec:abstract}
\noindent
We present the first cross-modality distillation framework specifically tailored for single-panoramic-camera Bird's-Eye-View (BEV) segmentation.
Our approach leverages a novel LiDAR image representation fused from range, intensity and ambient channels, together with a voxel-aligned view transformer that preserves spatial fidelity while enabling efficient BEV processing. 
During training, a high-capacity LiDAR and camera fusion Teacher network extracts both rich spatial and semantic features for cross-modality knowledge distillation into a lightweight Student network that relies solely on a single 360-degree panoramic camera image. Extensive experiments on the Dur360BEV dataset demonstrate that our teacher model significantly outperforms existing camera-based BEV segmentation methods, achieving a 25.6\% IoU improvement. Meanwhile, the distilled Student network attains competitive performance with an 8.5\% IoU gain and state-of-the-art inference speed of 31.2 FPS. 
Moreover, evaluations on KITTI-360 (two fisheye cameras) confirm that our distillation framework generalises to diverse camera setups, underscoring its feasibility and robustness.
This approach reduces sensor complexity and deployment costs while providing a practical solution for efficient, low-cost BEV segmentation in real-world autonomous driving.
The code is available at: \url{https://github.com/Tom-E-Durham/KD360-VoxelBEV}.
\end{abstract}
    
\vspace{-0.2cm}
\section{Introduction}
\label{sec:intro}

\begin{figure}[ht]
  \centering
  \includegraphics[width=1.0\columnwidth,]{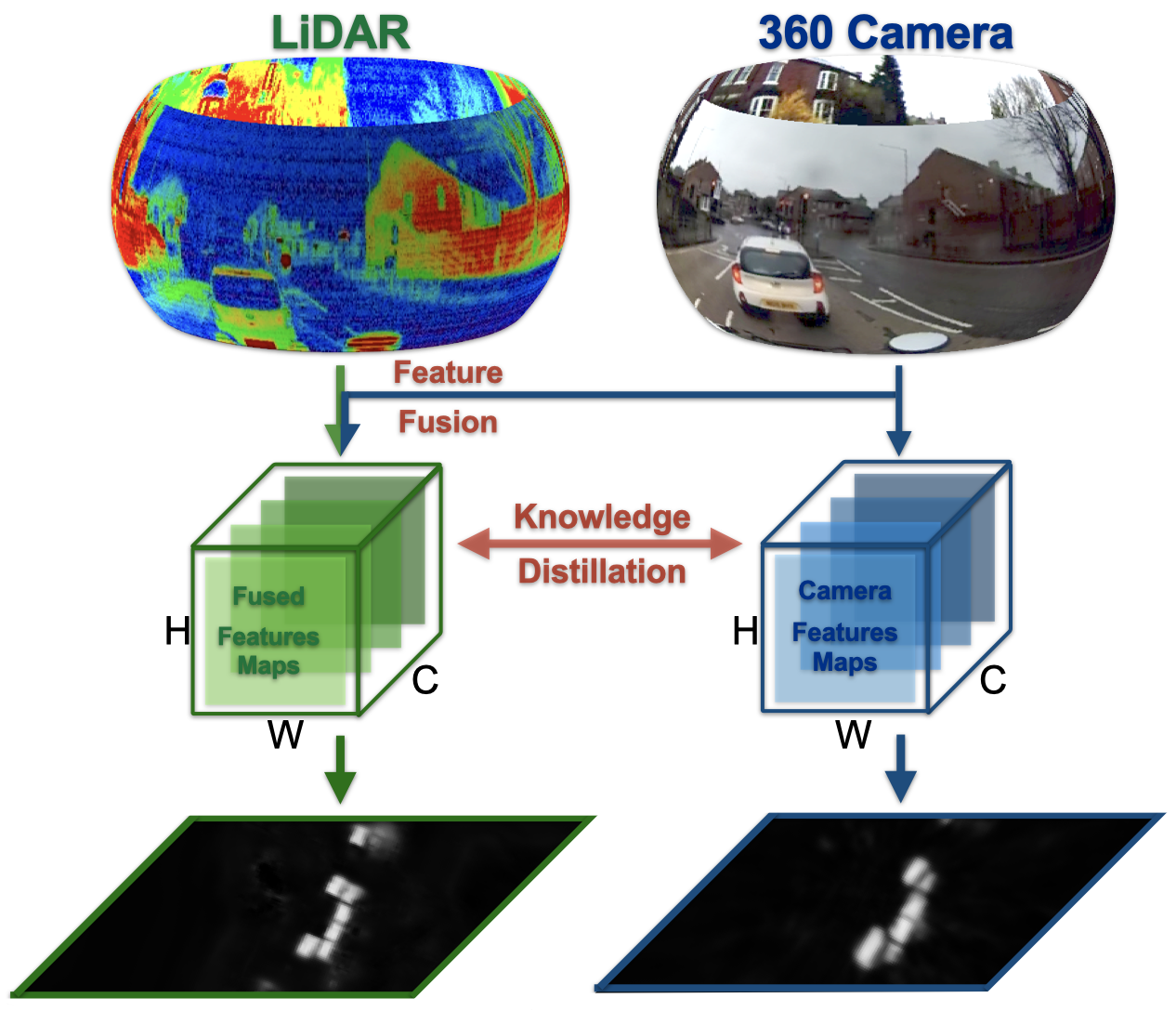}
  \caption{Overview of cross-modality channel-wise knowledge distillation from fused LiDAR–camera features (Teacher) to a single 360-degree camera model (Student) for enhanced feature representation and scene understanding.}
  \label{fig:overview}
  \vspace{-0.3cm}
\end{figure}

Bird's-Eye-View (BEV) representations play a crucial role in autonomous driving, offering a top-down perspective that greatly simplifies sensor fusion~\cite{man2023bev,ye2023fusionad,singh2023vision,li2023bevdepth} and benefits downstream tasks such as detection~\cite{bevformer2022,liu2023bevfusion,ye2025bevdiffuser}, segmentation~\cite{fiery2021,LSS2020,simplebev2023,pointbev2024,chabot2024gaussianbev}, motion forecasting~\cite{Ettinger_2021_ICCV, zhang2024simpl}, and navigation planning~\cite{hu2023planning,qin2023unifusion}. 
While most BEV generation approaches rely on multi-camera solutions ~\cite{simplebev2023,pointbev2024,chabot2024gaussianbev,bevformer2022,bevformerv22022}, the simultaneous deployment of multiple sensors introduces significant hardware complexity and cost. In contrast, a single 360-degree camera must distribute its finite number of pixels across the entire panorama, leading to lower per-pixel density and reduced feature quality ~\cite{dur360bev2025, teng2024360bev,qianSurveyFishEyeCameras2022}. This resolution constraint poses additional challenges for downstream tasks like BEV segmentation, which rely on precise object boundaries and detailed spatial cues.

Recent advances show that scaling backbone depth can mitigate low-resolution inputs by extracting richer features~\cite{simplebev2023,pointbev2024}. However, accuracy gains quickly plateau with model size, incurring prohibitive compute, latency, and on-vehicle energy costs for real-time deployment. In parallel, LiDAR supplies high-fidelity 3D measurements~\cite{raj2020survey} and multimodal fusion (camera+LiDAR/radar) attains strong performance~\cite{liu2023bevfusion,liang2022bevfusion}; yet keeping LiDAR active at inference introduces substantial cost, maintenance, and size/weight/power burdens, limiting large-scale feasibility. This raises a practical question: can we exploit LiDAR during training while discarding it at deployment?

We address this with \emph{KD360-VoxelBEV}, a cross-modality distillation framework that leverages LiDAR for training but relies only on a single camera at inference. 
Unlike existing cross-modality distillation works in BEV detection or HD map construction~\cite{klingner2023x3kd,ji2024vexkd,Distillbev2023,mapdistill,bevdistill2022}, our focus is on the under-explored single panoramic camera BEV segmentation setting, where the 360-degree equirectangular geometry and the low-resolution constraint introduce unique challenges that prior multi-camera methods do not address.
We introduce a high-capacity Teacher network that fuses LiDAR and camera features in BEV space via a soft-gated mechanism, while our LiDAR representation and voxel-aligned projection ensure that LiDAR-derived features retain both geometric fidelity and compatibility with 2D backbones. In contrast to existing approaches~\cite{simdistill,mapdistill} that enforce direct alignment across different modalities and often yield suboptimal results, our module employs \emph{Soft-Gated Fusion Module} (SGFM) to reduce representation discrepancies and further introduces an \emph{Auxiliary Module} (AM) to provide additional distillation signals, thereby facilitating faster convergence of the Student. These fused features supervise a lightweight camera-only Student, enabling deployment-friendly BEV segmentation without sacrificing accuracy.

We primarily develop and evaluate \emph{KD360-VoxelBEV} on Dur360BEV~\cite{dur360bev2025}, the latest real-world 360\degree{} single-camera dataset with the highest-resolution LiDAR among existing benchmarks. To further test generalisation, we also conduct experiments on KITTI-360, which provides two fisheye cameras with full 360-degree coverage.

\noindent The main contributions of this work are as follows:
\begin{itemize}[leftmargin=*]
\item We propose \emph{KD360-VoxelBEV}, the first cross-modality distillation framework for BEV segmentation, where the SGFM integrates LiDAR and camera features in the Teacher, providing strong multimodal supervision to a lightweight panoramic camera-only Student.
\item We introduce a novel voxel-aligned view transformer that preserves the geometric fidelity of LiDAR while remaining compatible with efficient 2D backbones, thereby combining the accuracy of voxel-based encoders~\cite{zhou2018voxelnet} with the efficiency of pillar-based designs~\cite{lang2019pointpillars}.
\item We design a unified panoramic LiDAR image representation, co-parameterised with equirectangular camera imagery, which ensures consistent cross-modal alignment and efficient 2D processing. In conjunction with the voxel-aligned view transformer, this representation enables a strong BEV Teacher and supports robust cross-dataset distillation.
\end{itemize}

\section{Related Work}
\label{sec:related_work}

We consider prior work in two related topic areas: BEV segmentation (Section \ref{subsec:bev_seg}) and cross-modality knowledge distillation (Section  \ref{subsec:cross-modality_KD}). 
\subsection{BEV Segmentation}
\label{subsec:bev_seg}
\noindent 
BEV segmentation maps provide a view of the environment for tasks such as scene understanding, path planning and motion prediction~\cite{hu2023planning}. Moreover, BEV representations fuse data from multiple sensors (e.g., LiDAR, camera, radar)~\cite{Qin_2023_ICCV, liu2023bevfusion, Zhao_2024_CVPR, sun2025tomd}. BEV methods differ mainly in their inputs: image-based approaches rely solely on camera data, while multi-modality methods integrate additional sensors.

\subsubsection{Image-Based vs. Multi-Modality BEV Approaches}
Most image-based BEV methods rely on multi-camera setups, typically employing six cameras, to provide a complete 360-degree surround view around the vehicle. This configuration offers high-definition views in all directions and provides dense semantic information. However, the multi-camera approach increases the memory requirements for data input and may introduce inter-view redundancy. In contrast, a single 360-degree camera input reduces the input data memory requirements while still capturing sufficient scene context. 
Other works fuse multiple sensors, such as LiDAR and radar, to enhance the performance of BEV segmentation by providing point cloud.
\vspace{-0.3cm}
\paragraph{Image-Based BEV:} 
Recent work includes LSS~\cite{LSS2020}, which employs an end-to-end pipeline that transforms multi-view camera images into a unified BEV representation by implicitly unprojecting 2D features to 3D space. 
FIERY~\cite{fiery2021} refines this by applying a multi-task framework with uncertainty weights and a probabilistic model for predicting future instance segmentation and motion in BEV.
BEVFormer~\cite{bevformer2022} and its updated version BEVFormerV2~\cite{bevformerv22022} adopt transformer-based architectures to effectively fuse multi-view inputs, demonstrating the potential of more complex models in generating rich BEV representations.
SimpleBEV~\cite{simplebev2023} introduces a parameter-free feature pulling method to replace the need for predicting depth distribution, whilst PointBEV~\cite{pointbev2024} presents a novel sparse paradigm to replace the dense grid sampling method, which offers flexible memory/performance trade-offs.
Furthermore, GaussianBEV~\cite{chabot2024gaussianbev} applies an optimisation free 3D Gaussian generator to transform image feature map into 3D Gaussians. Related monocular depth estimation work leverages semantic segmentation priors to regularise single-view 3D reconstruction~\cite{atapour2019segmentwisedepth}.
\vspace{-0.3cm}
\paragraph{Multi-Modality BEV:}
Several approaches improve BEV segmentation by incorporating additional sensors beyond cameras. For instance, SimpleBEV~\cite{simplebev2023} enhances performance by down-sampling radar and LiDAR data into grid representations, which are then concatenated with RGB features. Similarly, BEVFusion~\cite{liu2023bevfusion} uses a LiDAR encoder to extract features that are flattened along the $z$-axis before merging with RGB features in BEV space. While these fusion methods boost segmentation performance, they introduce extra processing steps that increase computational load and may slow inference, posing challenges for real-time deployment. Additionally, PointBEV~\cite{pointbev2024} departs from conventional fusion methods by proposing a coarse and fine training using flattened LiDAR data as sample points. These 2D BEV points then serve as sample locations for constructing 3D pillars of fixed height for sparse feature extraction. Nevertheless, the reliance on fixed-height pillars overlooks crucial vertical geometry, leading to a loss of LiDAR-specific spatial detail.

\subsection{Cross-Modality Knowledge Distillation}
\label{subsec:cross-modality_KD}
Traditional knowledge distillation is originally designed for transferring information from a large teacher model to a compact student model with the same input data~\cite{hinton2015distilling}, whereas cross-modality knowledge distillation supports distillation between different input modalities.
Previous cross-modality knowledge distillation model have been applied between data types: depth to RGB~\cite{Gupta_2016_CVPR}, sketch to photographic~\cite{han2017deepsketch2face}, and synthetic to real~\cite{Sankaranarayanan_2018_CVPR}. 
More recently, several works have adapted cross-modality distillation between LiDAR and cameras for 3D perception.
BEVDistill~\cite{bevdistill2022} introduces a dense foreground-guided feature imitation mechanism along with sparse instance-wise distillation to transfer rich spatial information from LiDAR to multi-camera 3D object detection networks.
Similarly, DistillBEV~\cite{Distillbev2023} employs region decomposition and adaptive scaling to achieve a more fine-grained cross-modal distillation, enhancing the alignment between LiDAR and camera modalities for BEV perception tasks.
UniDistill~\cite{UniDistill2023} adopts a three-level approach—feature, relation, and response—allowing for flexible teacher-student pairs of LiDAR or camera networks.

One challenge of cross-modality knowledge distillation is that it is not always effective when the representation and distribution gap between modalities is large, as forcing the student to mimic the teacher often leads to suboptimal performance~\cite{AM,c2kd}. Another challenge is that most existing approaches have focused on 3D object detection rather than BEV segmentation, and none address the 360-degree equirectangular format that naturally arises from a single panoramic camera. 
In this work, we bridge these gaps by proposing an SGFM-based cross-modality knowledge distillation framework with AM for BEV segmentation, introducing the first system that leverages LiDAR data to guide a single-camera model operating on full 360-degree equirectangular image input. 

Consequently, we do not provide direct quantitative comparisons to existing cross-modality distillation frameworks~\cite{ji2024vexkd,klingner2023x3kd,UniDistill2023,Distillbev2023,bevdistill2022,mapdistill} as the existing literature focuses on different tasks (e.g., 3D object detection or map reconstruction), employ LiDAR point cloud encoders and assume multi-camera rigs, while our framework targets single panoramic camera BEV segmentation using a unified LiDAR–camera equirectangular image representation.
\noindent

\section{Methodology}
\label{methodology}

Here, we introduce our novel LiDAR-driven distillation framework (Section~\ref{sec:methods:lidar-representation}), which includes a new LiDAR data representation, a voxel-aligned view transformer (Section~\ref{sec:Voxel-aligned View Transformer}), and a cross-modality knowledge distillation architecture (Section~\ref{sec:distill}).

\begin{figure}[htbp]
    \centering
    \begin{subfigure}{\linewidth}
        \centering
        \includegraphics[width=\linewidth, trim=1.5cm 1.5cm 1.5cm 1.5cm, clip]{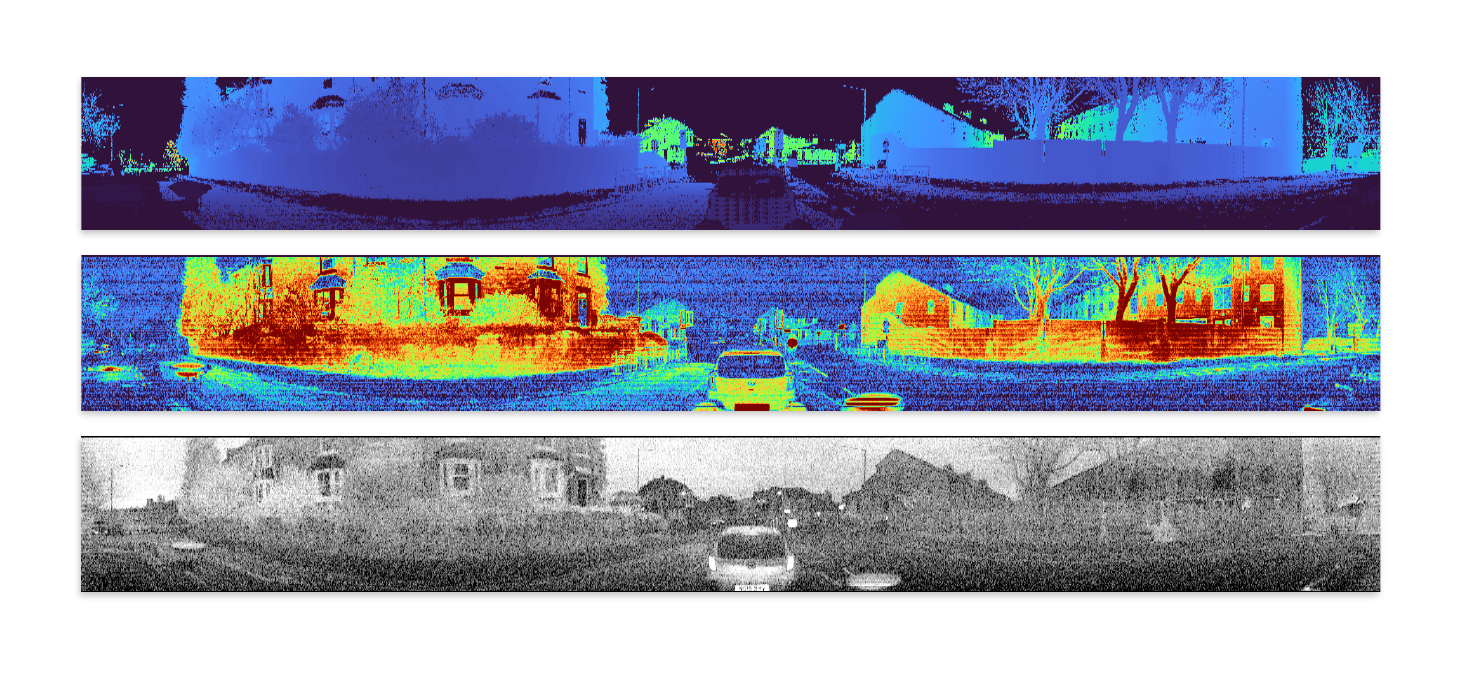}
        \caption{}
        \label{fig:dur360bev_lidar_img}
    \end{subfigure}

    \begin{subfigure}[b]{0.49\linewidth}
        \centering
        \includegraphics[width=\textwidth]{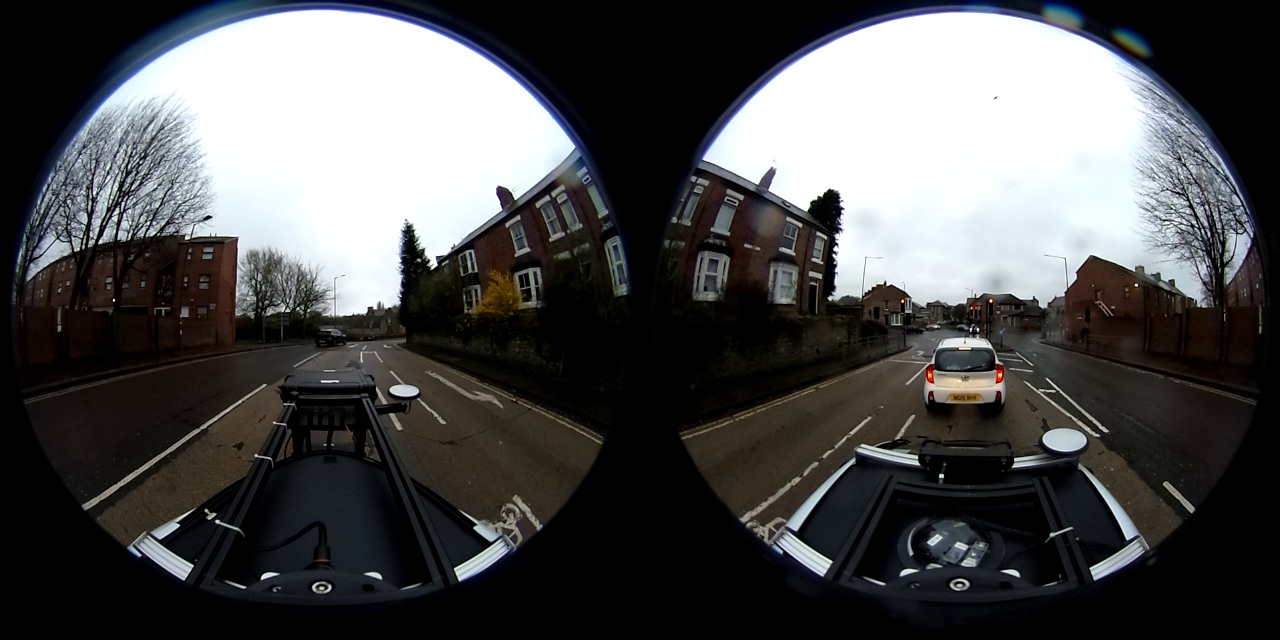}
        \caption{}
        \label{fig:dur360bev_df_img}
    \end{subfigure}
    \begin{subfigure}[b]{0.49\linewidth}
        \centering
        \includegraphics[width=\textwidth]{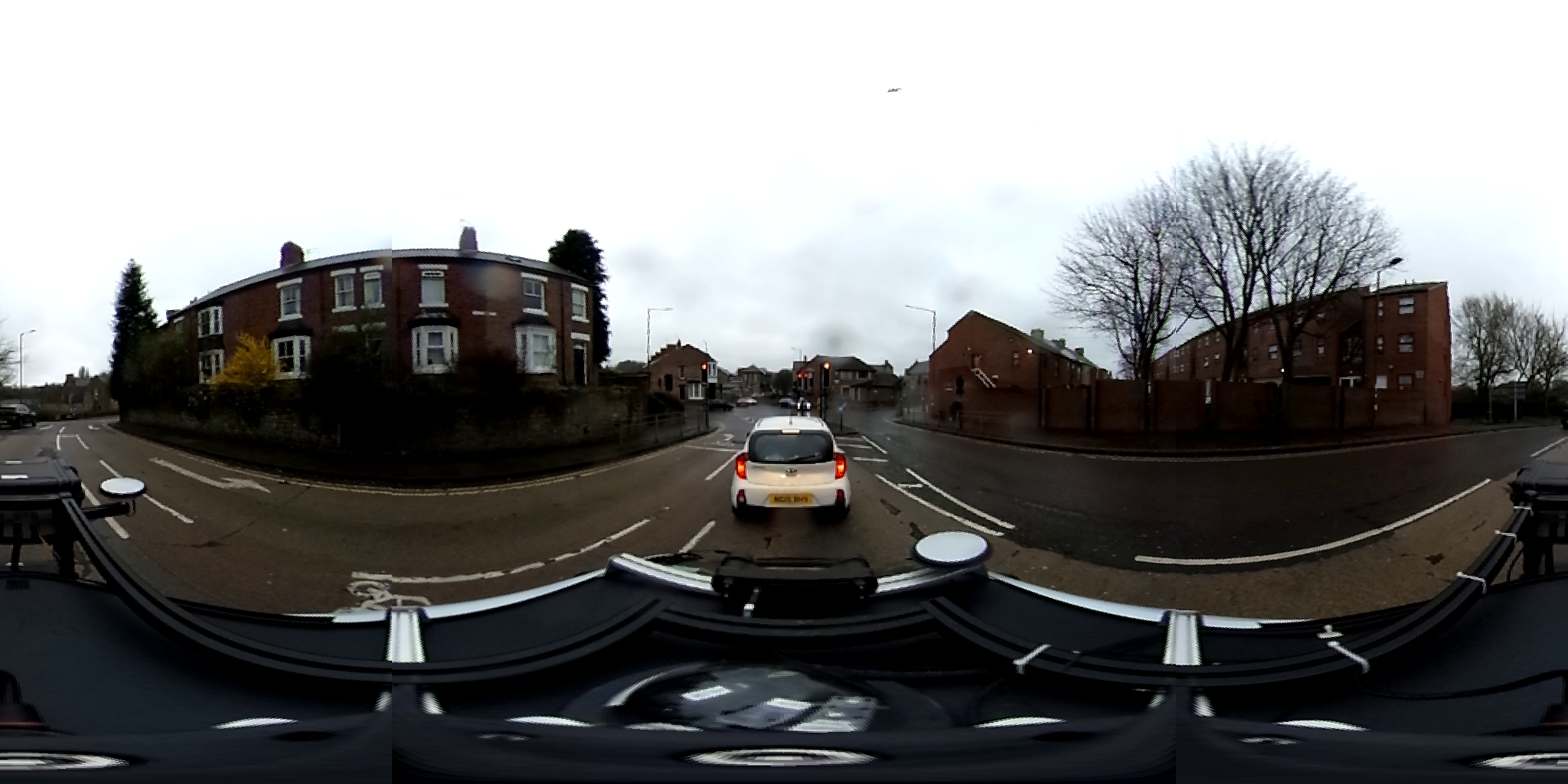}
        \caption{}
        \label{fig:dur360bev_equi_img}
    \end{subfigure}
    \vspace{-0.3cm}
    \caption{\textbf{Illustration of the Dur360BEV dataset~\cite{dur360bev2025}.} (a) LiDAR data in equirectangular representation [Top: range image; Middle: intensity image; Bottom: ambient image]. (b) Dual-fisheye spherical image. (c) Equirectangular-projected 360-degree image. }
    \label{fig:dur360bev_mainfigure}
\vspace{-0.3cm}
\end{figure}

\subsection{Unified LiDAR Image Representation}
\label{sec:methods:lidar-representation}
We introduce a unified LiDAR image representation that encodes raw point clouds into a dense equirectangular grid of shape \((H{\times}W{\times}3)\). Each pixel corresponds to a unique combination of azimuth and elevation for a full 360-degree coverage consistent with the sensor's field of view. The three channels store \emph{range}, \emph{intensity} and \emph{ambient}, thereby capturing both geometric and radiometric information. 
This representation offers two key advantages. First, it integrates seamlessly into efficient image-based BEV segmentation architectures, fully exploiting the detailed spatial and contextual information provided by high-resolution LiDAR.
Second, by expressing both LiDAR and RGB in a consistent spherical image space, it ensures natural alignment between modalities and thus reduces the distribution gap in cross-modality knowledge distillation (Figure~\ref{fig:dur360bev_mainfigure}).

Dur360BEV~\cite{dur360bev2025} natively provides all three fields for constructing this representation(as shown in Figure~\ref{fig:dur360bev_lidar_img}), whereas other datasets offer only a subset of modalities. The corresponding dataset-specific adjustments are described in Section~\ref{sec:experiments}.

While range and intensity are widely adopted in LiDAR-based perception frameworks, the ambient modality is less commonly explored. 
Unlike range, which encodes geometric distance, or intensity, which measures the strength of actively reflected laser pusles, ambient captures the background near-infrared radiation (800–2500 nm) received by the sensor. 
It therefore provides information complementary to range and intensity, reflecting scene properties under natural illumination conditions rather than active returns alone. 
This additional modality enriches our unified LiDAR image representation (Figure~\ref{fig:dur360bev_lidar_img}) and offers robustness in scenarios with challenging lighting, such as nocturnal or adverse-weather environments, without replacing the geometric or reflective cues already available.

\subsection{Voxel-Aligned View Transformer}\label{sec:Voxel-aligned View Transformer}

\begin{figure}
    \centering
    \includegraphics[width=1\linewidth]{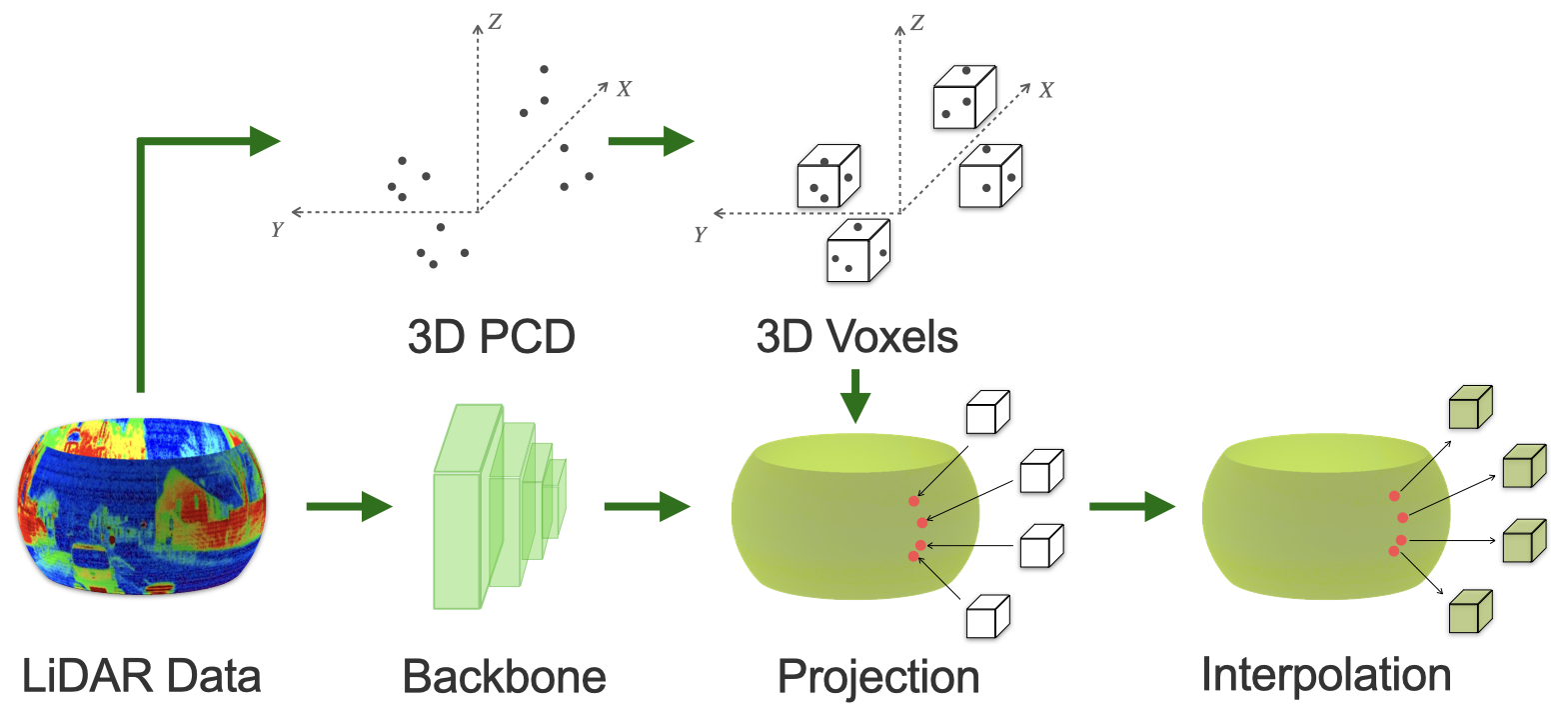}
    \caption{\textbf{Sparse Voxel Pulling Module (View Transformer).} 3D voxels derived from LiDAR point cloud are mapped to localised equirectangular LiDAR features, then bilinearly interpolated to form 3D BEV features.}
    \label{fig:voxel_pulling}
\vspace{-0.3cm}
\end{figure}    

We introduce a novel \textit{voxel-aligned view transformer}, a feature projection module that maps equirectangular image features into the BEV space guided by LiDAR voxelisation (as shown in Figure \ref{fig:voxel_pulling}). 
Unlike grid-based methods that predict on dense BEV grids~\cite{simplebev2023} or pillar-based approaches that form fixed-height vertical columns from 2D BEV points~\cite{pointbev2024}, our design aligns voxels directly with the underlying LiDAR geometry.
This yields a representation that combines the strengths of voxel- and pillar-based models. It preserves the spatial fidelity of voxel features while, once compressed into the BEV plane, remaining lightweight and directly amenable to standard 2D CNN backbones. In contrast to pillar-based methods that risk sampling outside the true object extent, our voxel alignment ensures precise and noise-free feature projection.

Formally, given an equirectangular feature map \(\mathbf{I} \in \mathbb{R}^{C{\times}H{\times}W}\) extracted from the input image backbone, we voxelise the LiDAR point cloud \(\mathbf{P} = \{ p \mid p = (x,y,z) \in \mathbb{R}^3 \}\) within a detection range \([-R,R]\) using voxel size \((r_x,r_y,r_z)\). 
Each voxel centre is then mapped to image coordinates, and bilinear interpolation retrieves the corresponding visual features, producing a voxel-level feature tensor \(\mathbf{F} \in \mathbb{R}^{B{\times}C{\times}Z{\times}Y {\times}X}\). 
This operation leverages the 360-degree coverage of the equirectangular format to fully utilise the voxel grid without suffering from multi-camera overlaps or out-of-FoV artefacts~\cite{simplebev2023,bevformer2022,bevformerv22022, pointbev2024}. 

Finally, the voxel features are compressed along the vertical axis to yield a BEV feature map, which is further refined using a lightweight sparse U-Net to generate segmentation predictions. 

In summary, our design combines the unified LiDAR image representation with the voxel-based view transformer. 
This integration fully exploits the geometric and radiometric data from LiDAR since the voxel projection preserves spatial accuracy, while the image-based representation ensures compatibility with efficient 2D CNN architectures. 

\begin{figure*}[ht]
  \centering
  \includegraphics[width=\textwidth]{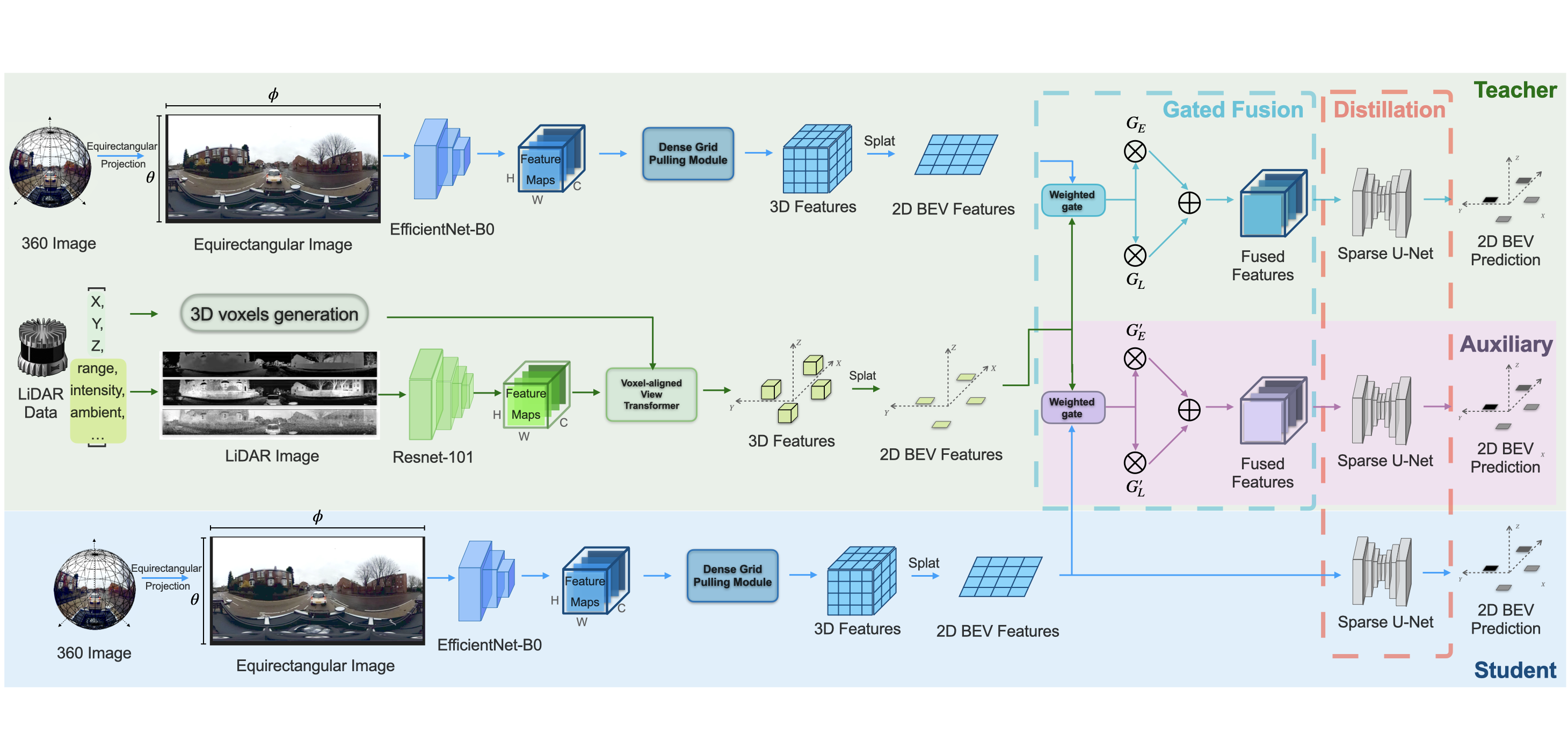}
  \vspace{-1cm}
  \caption{\textbf{Overview of the proposed \emph{KD360-VoxelBEV} architecture.} 
\textbf{Teacher network (green):} equipped with the \textbf{SGFM (blue dashed block)}, which integrates LiDAR range, intensity, and ambient cues with 360-degree camera features to produce enriched BEV representations. 
\textbf{AM (pink):} fuses the Student and pre-trained LiDAR branch during training to reduce the feature gap between Teacher and Student, providing additional reliable guidance. 
\textbf{Student network (blue):} a camera-only BEV segmentation model that benefits from cross-modal distillation, achieving robust BEV predictions from a single 360-degree input image. 
\textbf{Distillation (red dashed block):} highlights the regions where multi-channel dense feature distillation is applied, specifically between Teacher–Student and Student–Auxiliary pairs. At inference, only the Student network is employed, ensuring lightweight and deployment-friendly BEV segmentation.  
}
  \label{fig:architecture}
\vspace{-0.3cm}
\end{figure*}

\subsection{Knowledge Distillation}\label{sec:distill}
\paragraph{Overview.} As illustrated in Figure~\ref{fig:architecture}, our approach employs a soft-gated LiDAR–camera \emph{Teacher} model (Section~\ref{sec:teacher}) to guide a cost-effective \emph{Student} (Section~\ref{sec:student}) that operates solely on single-camera images. The \emph{Teacher} uses SGFM  (Section~\ref{sec:softgate}) to adaptively balance geometric information from LiDAR with appearance cues from the camera yielding richer multimodal representations. To further reduce the feature gap between \emph{Teacher} and \emph{Student}, we introduce an AM (Section~\ref{sec:am}) during training. These spatial and semantic insights are distilled into \emph{Student} to allow the camera-only model to achieve improved BEV segmentation accuracy at test time without the hardware and computational overhead of LiDAR sensors.

\subsubsection{Soft-Gated Fusion Module}\label{sec:softgate}
The proposed SGFM is designed to adaptively combine LiDAR- and image-derived BEV features. 
It consists of three main steps: feature concatenation, gate generation and gated fusion, followed by a refinement stage. 
Let $\mathbf{F}^{I} \in \mathbb{R}^{C{\times}H {\times}W}$ and $\mathbf{F}^{L} \in \mathbb{R}^{C {\times}H{\times}W}$ denote the feature maps from the image and LiDAR branches, respectively. 
After channel-wise concatenation, we obtain the joint feature map $\mathbf{F}^{cat} \in \mathbb{R}^{2C{{\times}}H{{\times}}W}$. 
A $1{{\times}}1$ convolution with weights $\mathbf{W} \in \mathbb{R}^{C{{\times}}2C{{\times}}1{{\times}}1}$ is applied to $\mathbf{F}^{cat}$ to model cross-modal correlations and predict a gating coefficient map:
\begin{equation}
\begin{aligned}
\mathbf{G}_{k,i,j} &= \sigma\Big( \sum_{k'=1}^{2C} \mathbf{F}^{cat}_{k',i,j}\,\mathbf{W}_{k,k'} \Big),\\
&\qquad \forall\, k \in [1,C],\; i \in [1,H],\; j \in [1,W].
\end{aligned}
\end{equation}
where $\sigma(\cdot)$ denotes the sigmoid function, ensuring $\mathbf{G} \in [0,1]^{C{{\times}}H{{\times}} W}$. 
Here, $\mathbf{G}$ acts as a soft gate: higher values increase the contribution of image features, while lower values emphasize LiDAR features. 
The gated fusion result is thus computed as
\begin{equation}
\mathbf{F}^{fuse} = \mathbf{G} \odot \mathbf{F}^{I} + (1 - \mathbf{G}) \odot \mathbf{F}^{L},
\end{equation}
where $\odot$ denotes element-wise multiplication. 
Finally, $\mathbf{F}^{fuse}$ is refined by a $3{{\times}}3$ convolution followed by batch normalization and ReLU activation:
\begin{equation}
\mathbf{F}^{out} = \phi \big( \mathrm{Conv}_{3{\times}3}([\mathbf{F}^{fuse}; \mathbf{F}^{fuse}]) \big),
\end{equation}
where $\phi(\cdot)$ denotes the BN–ReLU operation. 
This formulation allows the network to learn spatially varying, channel-aware weights that balance LiDAR geometry and image semantics, thereby yielding more reliable multimodal BEV representations for distillation.

\subsubsection{Teacher: Soft-Gated LiDAR-Camera Fusion}\label{sec:teacher} 
Our Teacher network consists of LiDAR and 360-degree camera branches whose features are fused through a soft-gated mechanism) in BEV space, providing a unified multimodal representation for supervision. The LiDAR inherently provides precise 3D geometric information, while the 360-degree camera contributes complementary visual appearance cues. Inspired by \cite{Cheng_2017_CVPR}, we adopt a late-fusion strategy at the BEV feature level, where modality-specific features are abstracted independently and subsequently fused after BEV conversion. 

Conventional approaches that directly operate on raw point clouds with dedicated 3D backbones followed by view transformation often suffer from distribution mismatch when combined with image features. To address this issue, we transform the LiDAR point cloud into a 3-channel equirectangular image representation, consisting of range, intensity, and ambient channels. This compact 2D representation enables the use of well-established convolutional backbones for effective feature extraction. The resulting feature maps are then projected into the BEV domain via our voxel-pulling module, which preserves fine-grained geometric details while maintaining global scene context. By combining these LiDAR features with those from the 360-degree image branch through a soft-gated fusion mechanism, the Teacher produces enriched BEV representations that serve as strong supervisory signals for distillation.

\subsubsection{Student: Camera-Only}\label{sec:student}
The Student model is lightweight and cost-effective, operating exclusively on a single 360-degree camera input. We adopt a compact 2D backbone to extract features and a dense view-transformation module (grid sampling) to map image features to BEV, mirroring the Teacher’s projection to ensure feature-level alignment during distillation. 
This design allows both networks to produce spatially compatible BEV features. 
Although the Student lacks direct 3D geometry and thus tends to produce coarser BEV features, we mitigate this limitation via cross-modality distillation that aligns the Student’s intermediate representations (and outputs) with those of the Teacher.

Following the Dur360BEV benchmark~\cite{duong2024robust}, the Student predicts three BEV heads: car segmentation, centerness, and offset. 
We supervise them with focal loss~\cite{ross2017focal} (segmentation), balanced MSE~\cite{ren2022balanced} (centerness), and $\ell_1$ loss (offset). 
These terms are combined into the Student’s multi-task objective with learned weights via uncertainty-based weighting~\cite{kendallMultiTaskLearningUsing2018}; see Eq.~\eqref{eq:student_task}.
\begin{equation}
\label{eq:student_task}
\mathbf{\mathcal{L}_{\text{stu}}}
= \lambda_1\,\mathcal{L}_{\text{seg}}
+ \lambda_2\,\mathcal{L}_{\text{cen}}
+ \lambda_3\,\mathcal{L}_{\text{off}}.
\end{equation}
\vspace{-0.4cm}
\subsubsection{Auxiliary Module}\label{sec:am}
\begin{figure}
    \centering
    \includegraphics[width=1\linewidth]{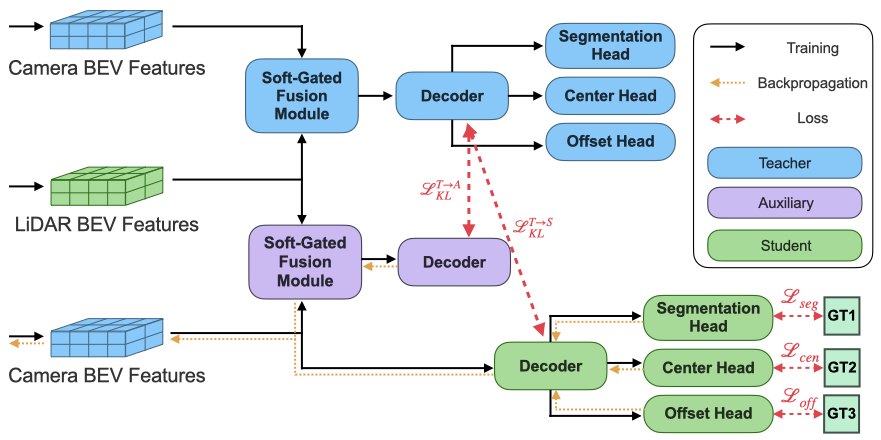}
    \caption{\textbf{Illustration of distillation and auxiliary details.} 
Feature maps from the decoder are used for channel-wise distillation, which is applied between Teacher and Student as well as between Auxiliary and Student, as indicated by the red dashed arrows.
}
    \label{fig:distill}
\end{figure}

The representation gap between the multimodal Teacher and the camera-only Student is substantial. Directly forcing the Student to match the Teacher often leads to unstable optimisation and suboptimal minima~\cite{AM}. 
To mitigate this, we introduce a training-only auxiliary branch that fuses the Student’s camera features with the Teacher’s LiDAR features using the same soft-gated mechanism. 
The auxiliary branch is supervised by the Teacher and provides an additional distillation signal to the Student, as illustrated in Figure~\ref{fig:distill}. 
This design supplies a smoother intermediate target that is closer to the Student’s modality while preserving LiDAR geometry for faster convergence. 
The auxiliary branch is disabled at inference, adding no runtime overhead.

\subsubsection{Channel-Wise Dense Feature Distillation}
Inspired by channel-wise knowledge distillation for dense prediction~\cite{shu2021channel}, we perform feature-level distillation at the decoder output (Fig.~\ref{fig:overview}). We investigate feature distillation at different network depths: early, middle and late stages, with detailed results reported in the ablation study (Section~\ref{sec:ablation}). We denote the fused BEV feature maps from the Teacher, Auxiliary, and Student as $\mathbf{F}^{T}$, $\mathbf{F}^{A}$, and $\mathbf{F}^{S}$ (each of size $C{{\times}}H{{\times}}W$). 
To compare features across modalities, we normalise each channel of the Teacher, Auxiliary, and Student feature maps into spatial probability distributions using a softmax with temperature $\mathcal{T}$. For channel $c$ this is defined as
\begin{equation}
\phi(\mathbf{F}_c) 
= \frac{\exp(\mathbf{F}_{c,i}/\mathcal{T})}
       {\sum_{i=1}^{H \cdot W} \exp(\mathbf{F}_{c,i}/\mathcal{T})}, 
\quad c = 1,\dots,C.
\end{equation}
where $i$ indexes spatial locations and $\mathcal{T}$ controls the distribution smoothness.  

The discrepancy between two normalised feature maps is measured using the Kullback–Leibler (KL) divergence. To aggregate across channels and compensate for temperature scaling, we define the KL-based distillation objectives for the Teacher–Student and Teacher–Auxiliary pairs as
\begin{equation}
\label{eq:kd_loss_ts}
\mathcal{L}_{\text{KL}}^{T\to S}
= \frac{\mathcal{T}^2}{C} \sum_{c=1}^{C} \sum_{i=1}^{H \cdot W} 
\phi(\mathbf{F}^T_{c,i}) \,\log \frac{\phi(\mathbf{F}^T_{c,i})}{\phi(\mathbf{F}^S_{c,i})},
\end{equation}

\begin{equation}
\label{eq:kd_loss_ta}
\mathcal{L}_{\text{KL}}^{T\to A}
= \frac{\mathcal{T}^2}{C} \sum_{c=1}^{C} \sum_{i=1}^{H \cdot W} 
\phi(\mathbf{F}^T_{c,i}) \,\log \frac{\phi(\mathbf{F}^T_{c,i})}{\phi(\mathbf{F}^A_{c,i})}.
\end{equation}
Ablation results analysing the effect of applying KL at different stages in Section~\ref{ab:stages}, 
and its advantage over alternative objectives in Section~\ref{ab:losses}.

The overall channel-wise distillation loss is then expressed as
\begin{equation}
\label{eq:kd_loss_full}
\mathcal{L}_{\text{KD}} 
= \alpha_{1}\,\mathcal{L}_{\text{KL}}^{T \to S} 
+ \alpha_{2}\,\mathcal{L}_{\text{KL}}^{T \to A},
\end{equation}

The overall training objective combines the channel-wise distillation loss in Eq.~\eqref{eq:kd_loss_full} and the multi-task Student loss in Eq.~\eqref{eq:student_task}:
\begin{equation}
\mathcal{L} = \mathcal{L}_{\text{KD}} + \mathcal{L}_{\text{stu}}.
\end{equation}


\vspace{-0.1cm}
\section{Experiments}
\label{sec:experiments}
We compare our approach with existing camera-based BEV segmentation methods. 
We benchmark against state-of-the-art methods on Dur360BEV~\cite{dur360bev2025} (Section~\ref{exp:dur360bev}) and further evaluate on KITTI-360~\cite{KITTI3602023} with a two-fisheye setup (Section~\ref{exp:kitti360}). 
We also present ablations (Section~\ref{sec:ablation}) on Dur360BEV to study distillation configurations and loss choices for \emph{KD360-VoxelBEV}.

\subsection{Datasets}
\textbf{Dur360BEV}~\cite{dur360bev2025} offers single panoramic-camera imagery paired with 128-channel LiDAR. 
We use the Extended version, featuring longer routes, more diverse scenes, and higher-quality annotations over the Initial release, with 13.9k frames for training and 1.5k for validation. 
Ground-truth BEV segmentation maps are derived from 3D bounding boxes, following the protocol of SimpleBEV~\cite{simplebev2023} and the official Dur360BEV setup~\cite{dur360bev2025}.
\textbf{KITTI-360}~\cite{KITTI3602023} provides two side-mounted fisheye cameras covering 360-degree and a 64-channel LiDAR. 
Since the dataset does not include BEV annotations, we generate ground-truth BEV segmentation maps following the same protocol as in Dur360BEV~\cite{dur360bev2025}(details in Appendix~\ref{supp:kitti360}). 
We adopt the split from~\cite{gosala2022bird} and downsample the size to 20k frames for training and 1k for validation due to computational constraint. 

We also considered other datasets, such as Mapillary~\cite{mapillary_metropolis_2021} and nuScenes~\cite{nuscenes2020}. 
The former provides panoramic imagery with fused point clouds but lacks per-frame LiDAR scans and dynamic objects, 
while the latter offers six surrounding-view cameras with 32-beam LiDAR but requires fusing the perspective images into an equirectangular view, introducing visible seams and artifacts~\cite{wei2024onebev}. 
These limitations make both datasets unsuitable for our BEV segmentation framework, see details in Appendix~\ref{supp:nus}.

\subsection{Evaluation Metrics}
We report Intersection over Union (IoU) across three square-shaped ranges: $100$m$\times$100$m$, $50$m$\times$50$m$, and $20$m$\times$20$m$. 
Following the Dur360BEV benchmark~\cite{dur360bev2025}, we also adopt the efficiency ratio (ER), defined as IoU divided by model parameters, to characterise the trade-off between accuracy and model complexity. In addition, we report frames per second (FPS) to measure inference speed.

\subsection{Experimental Setup}
In our \textit{voxel-aligned view transformer} module, we use voxels of size 0.5m in each dimension, and both the Teacher and the Student networks are configured within a detection range of $\left[100m\times8m\times100m\right]$ (in $x$,$y$,$z$ order, where $y$ is height) yielding a $200\times200$ BEV map covering $100m\times100m$ in the x–z plane. Following prior BEV segmentation works~\cite{simplebev2023,pointbev2024,dur360bev2025}, we adopt EfficientNet and ResNet as our backbones. Since our contribution focuses on the cross-modality distillation framework rather than backbone design, we intentionally use these widely-adopted baselines to ensure fair comparison and isolate the improvements brought by our distillation method.

Both Teacher and Student models are trained independently on an NVIDIA A100 80G GPU. For all main experiments, we train each network for up to 12k iterations with a batch size of 6, a learning rate of $5\times10^{-4}$, and a weight decay of $1\times10^{-5}$, using AdamW~\cite{loshchilov2017decoupled} with a 1-cycle learning rate schedule~\cite{smith2019super}. For inference, we evaluate all models on an NVIDIA RTX 3080 GPU and detailed configurations are provided in the Appendix~\ref{appendix:runtime}.



\subsection{State-of-the-Art Comparison}

\begin{figure*}[t]
    \centering
    \includegraphics[width=1.0\textwidth]{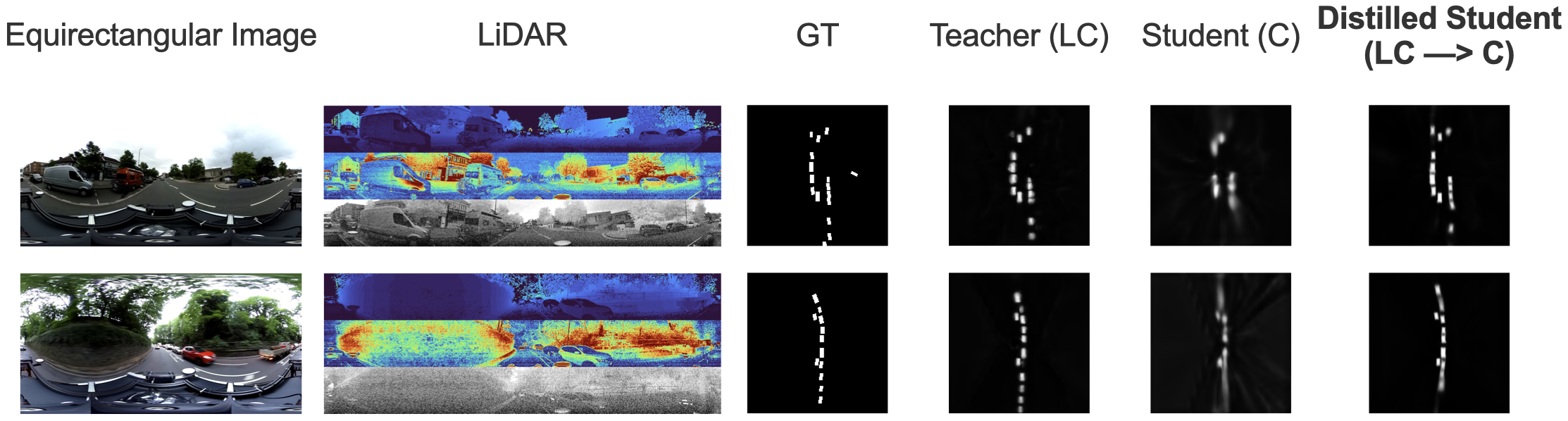}
    \caption{Qualitative BEV segmentation results on the Dur360BEV dataset~\cite{dur360bev2025}.}
    \label{fig:qualitative}
    
\end{figure*}

\paragraph{Dur360BEV.}
\label{exp:dur360bev}
Table~\ref{tab:dur360bev_experiments} reports a quantitative comparison with existing BEV segmentation methods, and the corresponding qualitative results are illustrated in Figure~\ref{fig:qualitative}. The comparison reveals three key findings: 

(i) By incorporating LiDAR through the proposed SGFM, the Teacher’s performance increases to 58.3\% $\text{IoU}_{100}$, confirming the effectiveness of multimodal fusion. The impact of different LiDAR channel combinations (Range, Ambient, and Intensity), as well as the effect of combining LiDAR with camera, are further analysed in the ablation study (see Appendix~\ref{supp:mdalities}). Results demonstrate that the configuration combining all three LiDAR channels with the camera achieves the best performance.

(ii) With the same lightweight EfficientNet-B0 backbone, our distillation framework improves the Student performance to 39.4\% $\text{IoU}_{100}$, surpassing all existing camera-only BEV segmentation methods. Moreover, the introduction of the AM further narrows the representation gap, pushing the distilled Student to 41.2\% $\text{IoU}_{100}$ while achieving the highest efficiency ratio. This demonstrates an excellent trade-off between accuracy and computational cost, making the model highly suitable for deployment in resource-constrained scenarios.


\paragraph{KITTI360.}
\label{exp:kitti360}
From Table~\ref{tab:kitti360_experiments}, since no prior methods are tailored for the KITTI-360 car segmentation task, we directly transfer our distillation configuration and evaluate it on this dataset. The results show that knowledge distillation continues to exploit multimodal advantages even under limited sensor settings, where LiDAR provides only two channels (range and intensity) and the cameras are wide-FoV fisheye. Compared with the Student baseline, our \emph{KD360-VoxelBEV} improves performance by +8.2\% $\text{IoU}_{100}$, demonstrating the robustness and generalisability of our approach across datasets with different modality characteristics.

\begin{table}[t]
  \centering
  \resizebox{\columnwidth}{!}{%
  \begin{tabular}{llllllll}
    \toprule
    Model & Modalities &Backbone & $\text{IoU}_{100}$ & $\text{IoU}_{50}$ & $\text{IoU}_{20}$ & ER & FPS\\
    \midrule
    SimpleBEV~\cite{simplebev2023}& C & RN-101 & 31.1 & 37.0 & 38.5 & 0.74 & \underline{25.7} \\
    PointBEV~\cite{pointbev2024}& C   & EN-b4  & 31.5 & 38.9 & 39.7 & 3.75 & 15.0\\
    \rowcolor{gray!10}
    Dur360BEV (dense)~\cite{dur360bev2025}& C & RN-101 & 32.7 & 40.4 & 42.0  & 0.78 & 25.3\\
    \rowcolor{gray!10}
    Dur360BEV (coarse/fine)~\cite{dur360bev2025}& C & EN-b4 & 32.6 & 40.3 & 41.6 &  \underline{3.88} & 14.9\\
    Ours (Teacher)& LC  & RN-101/EN-b0 & \textbf{58.3} & \textbf{64.3} & \textbf{68.0} & 0.85 &14.0\\
     Ours (Student)& C  & EN-b0 & 32.2 & 41.8 & 58.4 & 3.04 &\textbf{31.2}\\
    \rowcolor{gray!10}
    \textbf{KD360-VoxelBEV, distill w/o AM}& LC $\to$ C & EN-b0  & 39.4 & 49.3 & 64.2 & 3.72  &\textbf{31.2} \\
    \textbf{KD360-VoxelBEV, distill w/ AM}& LC $\to$ C & EN-b0  & \underline{41.2} & \underline{51.7} & \underline{67.1} & \textbf{3.89}  &\textbf{31.2}\\
    \bottomrule
  \end{tabular}}
  \addtolength{\tabcolsep}{1pt}
  \caption{Comparison results (\%) on \textbf{Dur360BEV} dataset~\cite{dur360bev2025} with metrics: IoU  $(\uparrow)$, ER $(\uparrow)$ and FPS $(\uparrow)$. \textbf{Bold} and \underline{underlined} numbers denote the best (Top-1) and second best (Top-2), respectively.
}
  \label{tab:dur360bev_experiments}
\end{table}

\begin{table}[t]
  \centering
  \resizebox{\columnwidth}{!}{%
  \begin{tabular}{llllll}
    \toprule
    Model & Modalities & Backbone & $\text{IoU}_{100}$ & $\text{IoU}_{50}$ & $\text{IoU}_{20}$ \\
    \midrule
    Ours (Teacher) & LC   & RN-101/EN-b0 & 53.9 & 65.8 & 76.4\\
     Ours (Student) & C   & EN-b0 & 24.1 & 37.9 & 63.8 \\
    \rowcolor{gray!10}
    \textbf{KD360-VoxelBEV}& LC $\to$ C & EN-b0  & 32.3 & 46.6 & 69.6 \\
    \bottomrule
  \end{tabular}}
  \addtolength{\tabcolsep}{1pt}
  \caption{Comparison results (\%) on \textbf{KITTI-360} dataset~\cite{KITTI3602023}.}
  \label{tab:kitti360_experiments}
  \vspace{-0.5cm}
\end{table}

\subsection{Ablation Studies}\label{sec:ablation}
\begin{table}[t]
  \centering
  \resizebox{\columnwidth}{!}{%
  \begin{tabular}{lllllll}
    \toprule
    Stage1 & Stage2 & Stage3 & AM & $\text{IoU}_{100}$ & $\text{IoU}_{50}$ & $\text{IoU}_{20}$ \\
    \midrule
    \checkmark & -- & -- & w/o & 20.96 & 31.53 & 47.05 \\
     -- & \checkmark & -- & w/o & 20.58 & 30.88 & 49.48 \\
     -- & -- & \checkmark & w/o & 39.37 & 49.34 & 64.21 \\
    \rowcolor{gray!10}
    \checkmark & -- & -- & w/ & 30.50 & 40.33 & 51.33 \\
    \rowcolor{gray!10}
     -- & \checkmark & -- & w/ & 32.22 & 42.11 & 56.42 \\
    \rowcolor{gray!10}
     -- & -- & \checkmark & w/ & 41.24 & 51.69 & 67.08 \\
    \bottomrule
  \end{tabular}
  }
 \addtolength{\tabcolsep}{1pt}
  \caption{Ablation study of KD model with different stage setups. 
 }
  \label{tab:stage_ablation}
  \vspace{-0.5cm}
\end{table}
We conduct a series of ablation experiments to better understand the factors driving the performance of \emph{KD360-VoxelBEV}.

\subsubsection{Effect of Distillation Stages and AM:}\label{ab:stages}
\begin{figure}
    \centering
    \includegraphics[width=0.8\linewidth]{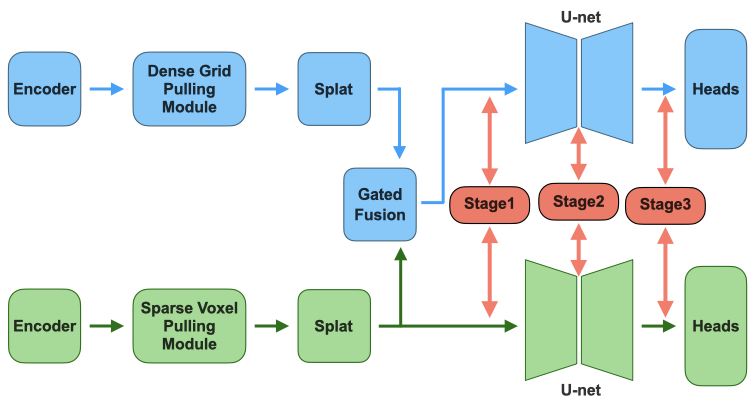}
    \caption{Illustration of distillation stages for ablation study (Auxiliary module excluded). Blue blocks denote Teacher modules, and green blocks denote Student modules.
} 
    \label{fig:stages}
    \vspace{-0.5cm}
\end{figure}

To evaluate the effectiveness of knowledge distillation, we consider three distinct stages (shown in Figure~\ref{fig:stages}): (i) after feature fusion (stage1; or, without gated fusion, the corresponding location in the LiDAR branch), (ii) the feature map at the output of the U-Net encoder (stage2), and (iii) the feature map at the output of the U-Net decoder (stage3). At each stage, we apply the channel-wise distillation loss in Eq.~\eqref{eq:kd_loss_ts} between Teacher and Student, optionally combined with the auxiliary distillation loss in Eq.~\eqref{eq:kd_loss_ta}. 

Overall, as shown in Table~\ref{tab:stage_ablation}, applying distillation at later stages yields larger performance gains, since early convolutional features are highly modality-specific and may introduce noise or instability when distilled, whereas deeper semantic representations are more modality-invariant and thus transfer more effectively across modalities. In addition, incorporating the AM further mitigates the representation gap between Teacher and Student, leading to consistent improvements across all IoU thresholds.


\subsubsection{Distillation Losses:}\label{ab:losses}
\begin{figure}
    \centering
    \includegraphics[width=0.7\linewidth]{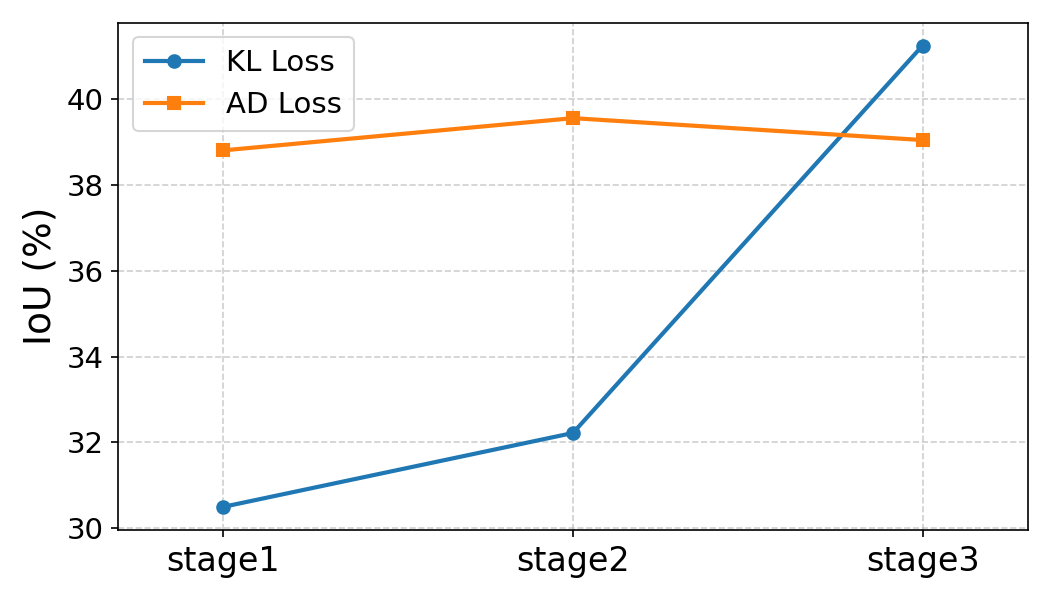}
    \caption{Comparison of distillation losses at different stages.} 
 \label{fig:losses}
 \vspace{-0.5cm}
\end{figure}

We evaluate two different distillation objectives: the Kullback–Leibler (KL) divergence loss~\cite{hinton2015distilling} and the Affinity Distillation (AD) loss~\cite{ad_loss}. A comparison of their performance is presented in Fig.~\ref{fig:losses}. As shown, the AD loss yields more stable performance across all stages, whereas the KL loss underperforms AD in the earlier stages but catches up at deeper layers. This trend can be attributed to the fact that KL divergence is more effective at aligning high-level semantic distributions than low-level spatial correlations.


\section{Conclusion}
\label{sec:conclusion}
We present KD360-VoxelBEV, a cross-modality distillation framework tailored to BEV segmentation from a single panoramic camera, where a soft-gated LiDAR–camera fusion Teacher supervises a lightweight camera-only Student. Our framework combines a voxel-aligned view transformer with a unified panoramic LiDAR image representation, preserving geometric fidelity while allowing for efficient BEV reasoning with 2D backbones. Experiments on Dur360BEV~\cite{dur360bev2025} and KITTI-360~\cite{KITTI3602023} show that the Teacher substantially outperforms existing camera-based BEV segmentation methods and that the distilled Student matches or surpasses state-of-the-art camera-only baselines~\cite{dur360bev2025,pointbev2024,simplebev2023} while running in real time. In future work, we plan to validate the framework on a broader range of datasets and investigate its integration with transformer-based architectures to further enhance representation capacity.

{
    \small
    \bibliographystyle{ieeenat_fullname}
    \bibliography{main}
}
\clearpage            
\appendix             
\clearpage
\begin{center}
  \LARGE \textbf{Appendix}\\[4pt]
  \large
\end{center}

\section{Dur360BEV-mini}\label{supp:mdalities}
Similar to the official nuScenes-mini~\cite{nuscenes2020} split (1\% of the full dataset, 10 scenes in total), 
we construct a mini version of Dur360BEV-Extended to facilitate efficient input modality ablation studies and debugging. 
Specifically, we uniformly downsample the Extended split by a factor of 10, 
resulting in 1,350 training and 150 validation frames. 
This reduced set preserves the distribution of the original dataset while enabling much faster experimentation. 
Unless otherwise stated, all reported main results are obtained on the full Dur360BEV-Extended dataset.

\section{nuScenes~\cite{nuscenes2020} Expriments}\label{supp:nus}
nuScenes~\cite{nuscenes2020} provides six surrounding-view cameras and a 32-channel LiDAR. 
We adopt the official train/val split (28.1k/6k samples) and generate BEV ground truth following the protocol in Lift-Splat~\cite{LSS2020} and SimpleBEV~\cite{simplebev2023}, where points inside “vehicle” bounding boxes are labeled positive and others negative.

\begin{figure}[htbp]
    \centering
    \begin{subfigure}{\linewidth}
        \centering
        \includegraphics[width=\linewidth]{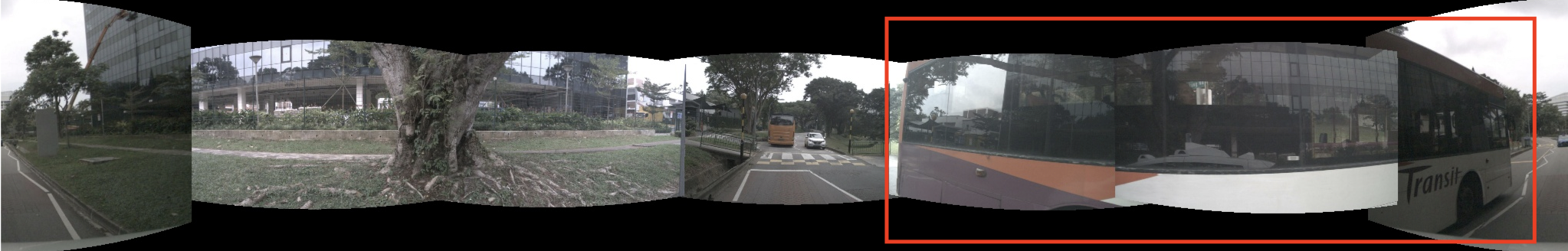}
        \caption{}
        \label{fig:nus_input}
    \end{subfigure}

    \begin{subfigure}[b]{0.25\linewidth}
        \centering
        \includegraphics[width=\textwidth]{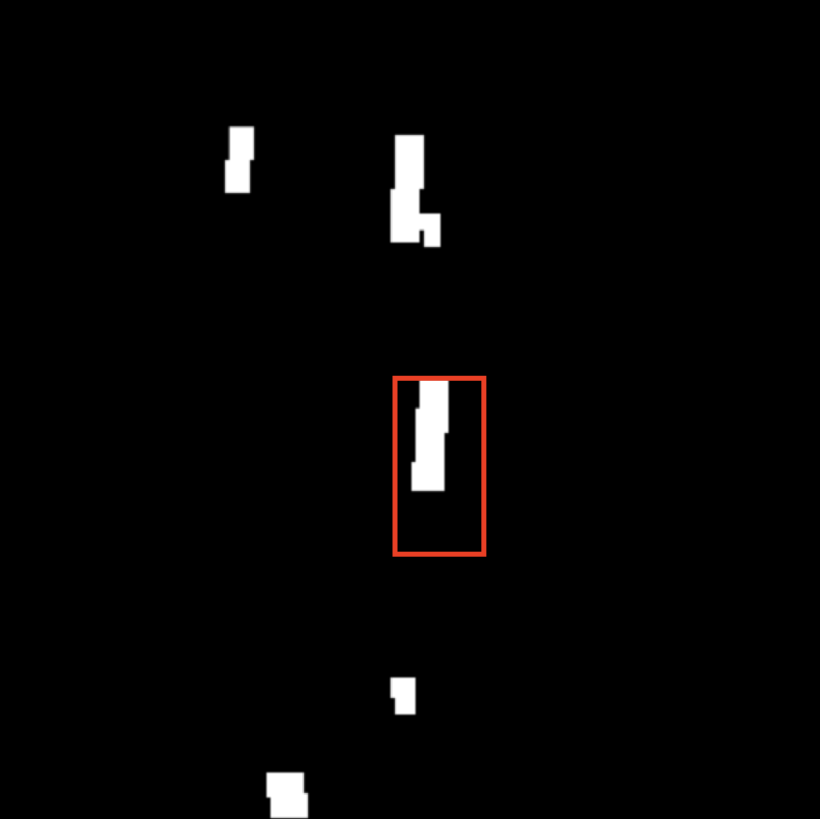}
        \caption{}
        \label{fig:nus_gt}
    \end{subfigure}
    \begin{subfigure}[b]{0.25\linewidth}
        \centering
        \includegraphics[width=\textwidth]{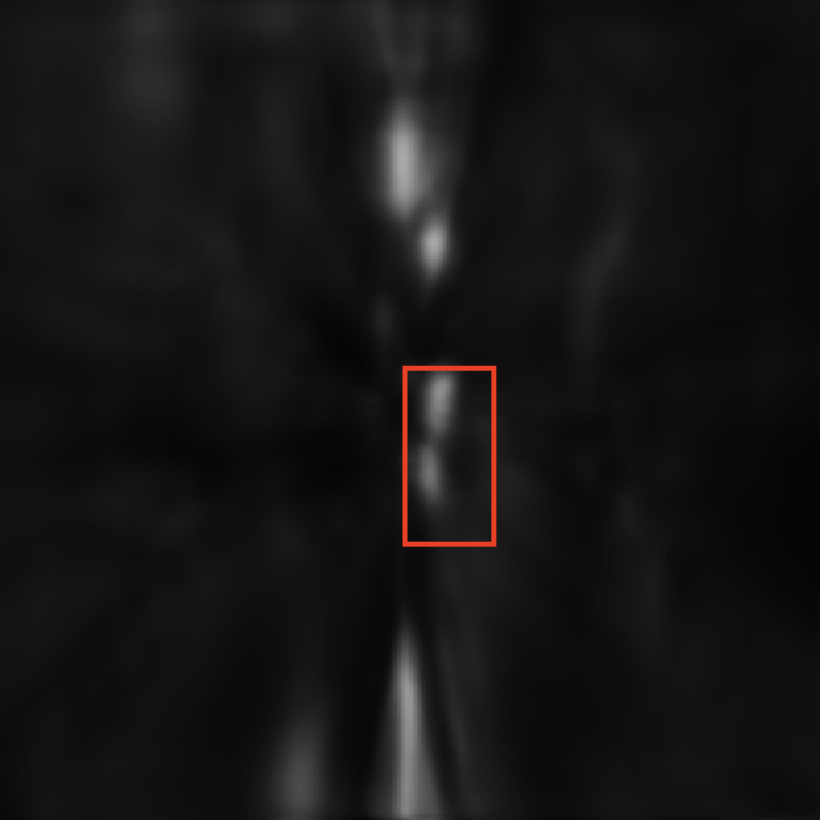}
        \caption{}
        \label{fig:nus_pred}
    \end{subfigure}
    \caption{\textbf{An illustrative case of suboptimal results when applying \emph{KD360-VoxelBEV} on nuScenes~\cite{nuscenes2020}.} (a) Equiangular image converted from six cameras; (b) BEV car label; (c) distillation segmentation results. The red box highlights a bus, which is fragmented due to discontinuities between stitched camera views.}
    \label{fig:nus_bad_case}
    \vspace{-0.5cm}

\end{figure}

We found that applying \emph{KD360-VoxelBEV} to nuScenes~\cite{nuscenes2020} is problematic due to the image formation process. As shown in Fig.~\ref{fig:nus_input}, the equiangular projection is stitched from six individual cameras rather than captured by a true 360\degree{} sensor. Consequently, objects spanning across multiple cameras often become misaligned at the image borders, leading to duplicated or fragmented appearances. In addition, differences in exposure and illumination across cameras introduce visible seams and inconsistencies. These artifacts make it difficult for the model to extract coherent features, and as illustrated by the prediction in Fig.~\ref{fig:nus_pred}), a single car may be broken into several disconnected parts. This highlights the limitation of using stitched multi-camera images in our distillation model, which relies on consistent 360\degree{} visual input.

\section{Effect of Input Modalities and SGFM}
\label{supp:modalities}

To enable faster experimentation, we perform this study on Dur360BEV-mini, while all main results are reported on the full dataset. 
The details of the Dur360BEV-mini are provided in Appendix~\ref{supp:mdalities}. We follow the same experimental configuration as in the main experiments, but train on the Dur360BEV-mini dataset up to 4k iterations to account for the reduced amount of input data.

\vspace{0.1cm}
\noindent\textbf{Impact of LiDAR Channels.} 
We first analyse the contribution of the three LiDAR channels (range, intensity, ambient) by selectively enabling or disabling them.
As shown in Table~\ref{tab:modality}, ambient information proves highly important: removing it (LiDAR (RI)) causes a notable performance drop compared to the full configuration (LiDAR (RAI)), decreasing $\text{IoU}_{100}$ from 54.0\% to 50.9\%.
The inclusion of ambient data consistently enhances BEV segmentation alongside range and intensity. This complementary cue, which captures environmental illumination beyond geometry and reflectivity, may explain why our distillation framework achieves stronger gains on Dur360BEV~\cite{dur360bev2025} compared to KITTI-360~\cite{KITTI3602023}, which lacks ambient measurements.

\vspace{0.1cm}
\noindent\textbf{Effectiveness of SGFM.}
We further validate the effectiveness of our Soft-Gated Fusion Module (SGFM) by comparing single-modality baselines against the fused system.
As reported in Table~\ref{tab:modality}, the Camera-only (C) baseline achieves 31.6\% $\text{IoU}_{100}$, limited by the lack of explicit depth information. The LiDAR-only (RAI) model performs significantly better at 54.0\% $\text{IoU}_{100}$ due to precise geometric sensing.
However, by integrating both modalities via SGFM, our Teacher model (LiDAR (RAI) + C) reaches 58.8\% $\text{IoU}_{100}$, outperforming the strongest single-modality baseline by +4.8\%.
This improvement indicates that SGFM effectively leverages the complementary nature of the two sensors—combining the rich semantic and texture cues from the $360^{\circ}$ camera with the accurate spatial geometry from LiDAR—to produce a more robust BEV representation.

\begin{table}[t]
  \centering
  \resizebox{\columnwidth}{!}{%
  \begin{tabular}{lccc}
    \toprule
    Input Modality & $\text{IoU}_{100}$ & $\text{IoU}_{50}$ & $\text{IoU}_{20}$ \\
    \midrule
    \textit{Single-Modality Baselines} & & & \\
    C (Camera Only)          & 31.6 & 37.9 & 39.6  \\
    LiDAR (RI)               & 50.9 & 56.8 & 62.0 \\
    LiDAR (AI)               & 51.7 & 58.5 & 67.0 \\
    LiDAR (RA)               & 53.4 & 59.2 & 65.1 \\
    LiDAR (RAI)              & 54.0 & 61.0 & 67.2 \\
    \midrule
    \textit{Multi-Modality Fusion} & & & \\
    \textbf{LiDAR (RAI) + C (w/ SGFM)} & \textbf{58.8} & \textbf{63.4} & \textbf{70.1} \\
    \bottomrule
  \end{tabular}}
  \caption{Ablation study on input modalities and SGFM using the Dur360BEV-mini~\cite{dur360bev2025}. Metrics: IoU $(\uparrow)$. \textbf{R}: Range, \textbf{A}: Ambient, \textbf{I}: Intensity, \textbf{C}: $360^{\circ}$ Camera. The proposed SGFM effectively fuses LiDAR and Camera features to outperform unimodal baselines.}
  \label{tab:modality}
  \vspace{-0.5cm}
\end{table}

\section{Inference Time Measurement Details}
\label{appendix:runtime}
We benchmark inference time for all models on the Dur360BEV dataset using a single NVIDIA RTX 3080 GPU. 
All measurements are conducted at a fixed input resolution of $1024 \times 2048$ under PyTorch with CUDA/cuDNN enabled. 
Each experiment is repeated twenty times, and the average is reported. 

Table~\ref{tab:runtime} summarizes the latency, throughput (FPS), and batch size used for inference. While the teacher model exhibits the slowest inference, KD360-VoxelBEV achieves the best trade-off between speed and efficiency, significantly surpassing existing methods and demonstrating both the effectiveness of knowledge distillation and its suitability for real-world applications.  

\begin{table}[h]
  \centering
  \resizebox{\columnwidth}{!}{%
  \begin{tabular}{lcc}
    \toprule
    Model & Latency (ms) $\downarrow$ & FPS $\uparrow$ \\
    \midrule
    SimpleBEV~\cite{simplebev2023}& 38.9 & 25.7 \\
    PointBEV~\cite{pointbev2024} & 66.7 & 15.0\\
    Dur360BEV (dense)~\cite{dur360bev2025} &38.5  & 25.3\\
    Dur360BEV (coarse/fine)~\cite{dur360bev2025} &67.1&14.9\\
    Ours (Teacher) & 71.7 & 14.0\\
    \textbf{KD360-VoxelBEV} & 32.1 & 31.2\\
    \bottomrule
  \end{tabular}%
  }
  \caption{Inference time comparison of different models on the Dur360BEV dataset using an NVIDIA RTX 3080 GPU.}
  \label{tab:runtime}
  \vspace{-0.5cm}
\end{table}

\section{KITTI-360}
\label{supp:kitti360}
\subsection{Image Processing}
\begin{figure}[htbp]
    \centering
    \begin{subfigure}{\linewidth}
        \centering
        \includegraphics[width=\linewidth]{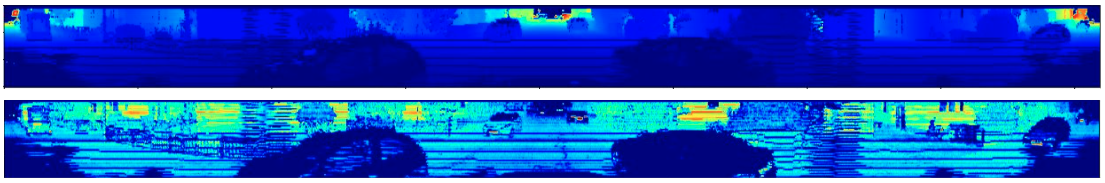}
        \caption{}
        \label{fig:kitti360_lidar_img}
    \end{subfigure}

    \begin{subfigure}[b]{0.49\linewidth}
        \centering
        \includegraphics[width=\textwidth]{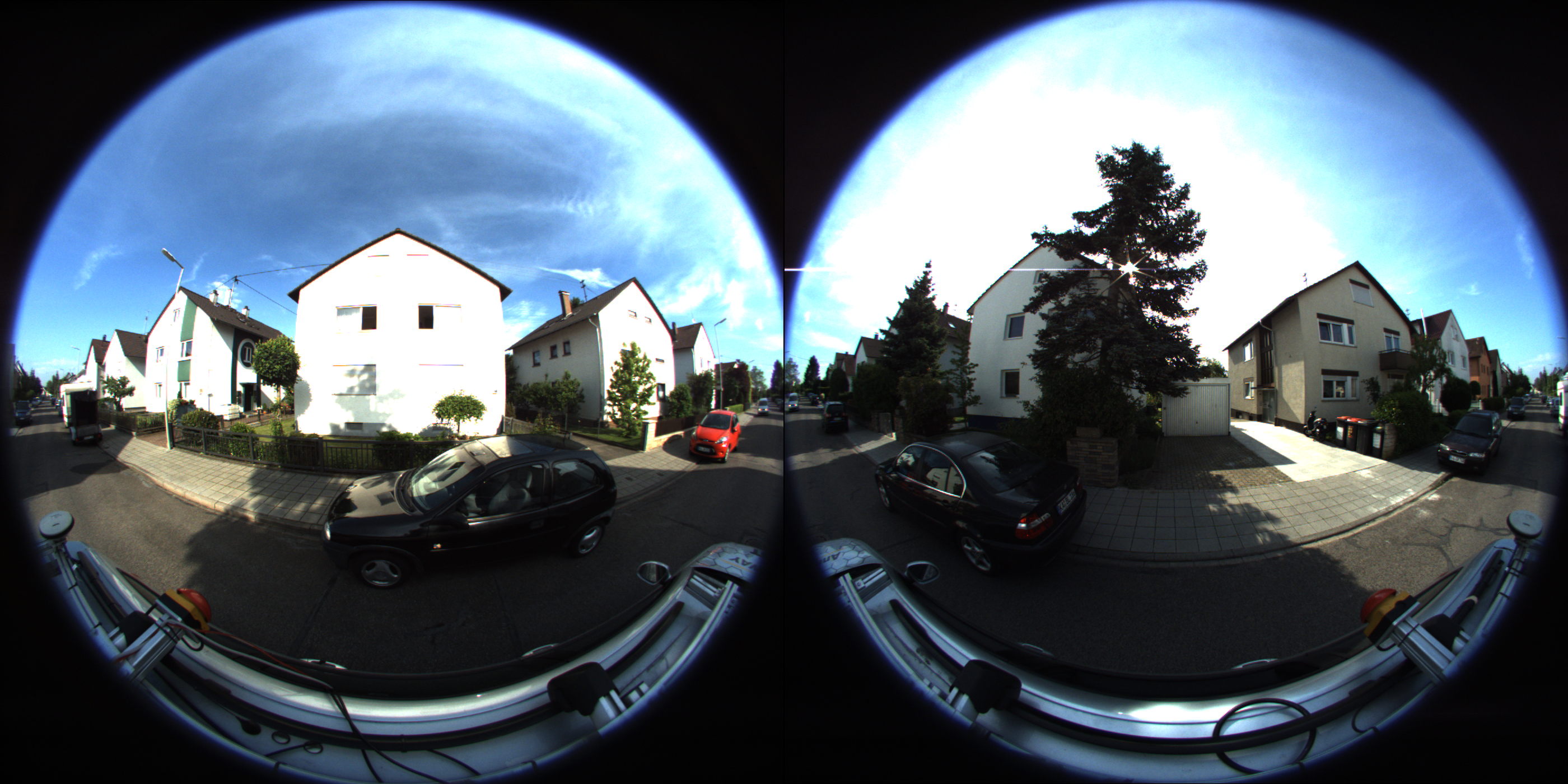}
        \caption{}
        \label{fig:kitti360_df_img}
    \end{subfigure}
    \begin{subfigure}[b]{0.49\linewidth}
        \centering
        \includegraphics[width=\textwidth]{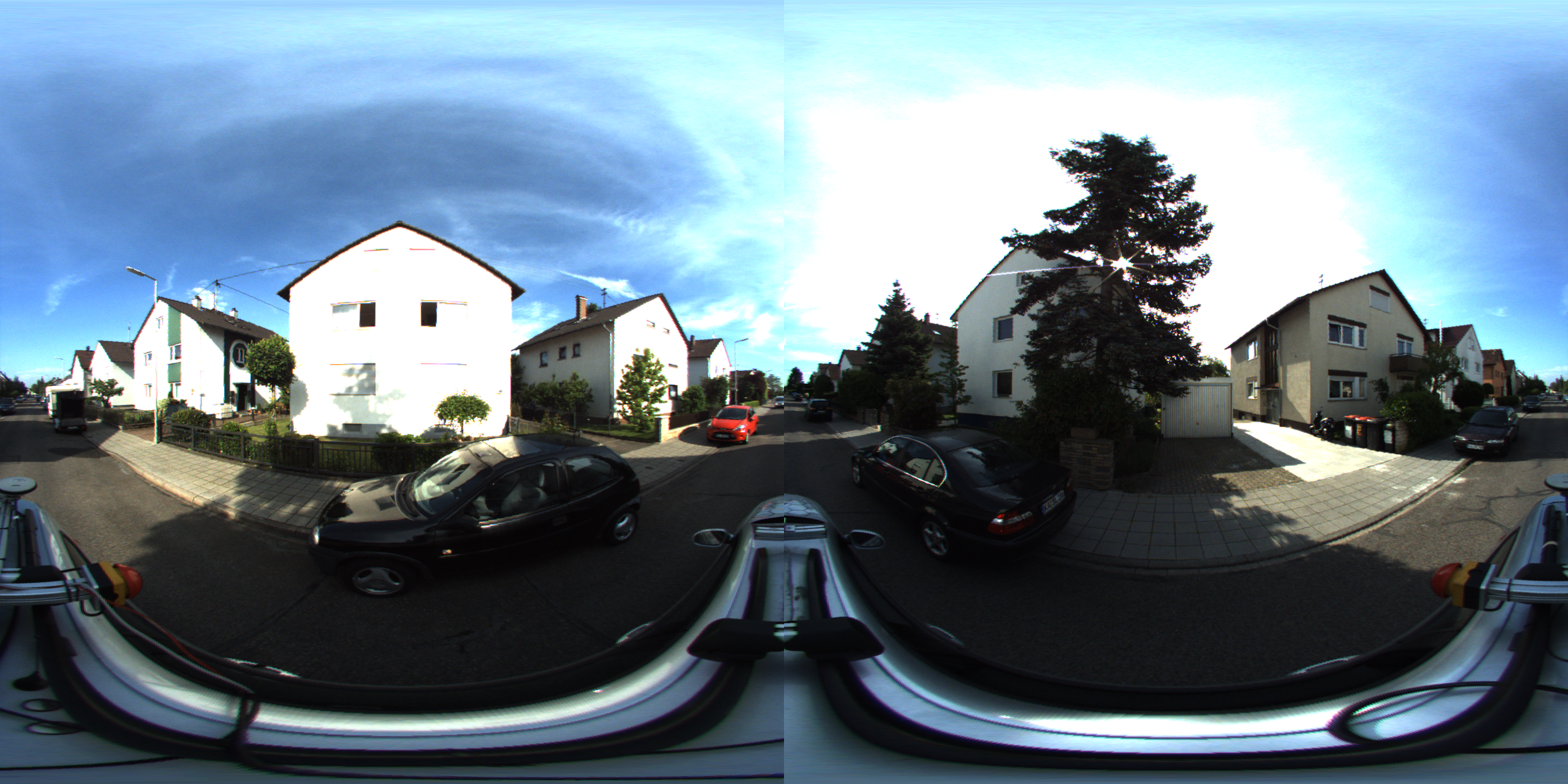}
        \caption{}
        \label{fig:kitti360_equi_img}
    \end{subfigure}
    \caption{\textbf{Illustration of the KITTI-360 dataset~\cite{dur360bev2025}.} (a) LiDAR data in equirectangular representation [Up: range image; Down: intensity image]. (b) Dual-fisheye spherical image. (c) Equirectangular-projected 360-degree image. }
    \label{fig:kitti360_dataset}
    \vspace{-0.5cm}
\end{figure}

KITTI-360 provides raw images from two side-mounted wide-FoV fisheye cameras. 
We project each fisheye image into spherical coordinates using the official calibration parameters and camera FoV, and then reproject to an equirectangular format. 
In this space, pixels are parameterised by azimuth and elevation angles, covering the full 360-degree field of view (see Figure~\ref{fig:kitti360_df_img}, \ref{fig:kitti360_equi_img}). 
This conversion yields a panoramic representation consistent with Dur360BEV, facilitating cross-dataset training and evaluation.

\subsection{Annotation Process}
The raw KITTI-360~\cite{KITTI3602023} annotations consist of two categories of 3D bounding boxes: static and dynamic objects. Static objects are labeled on accumulated point clouds, while dynamic objects are annotated on individual frames. Directly using these annotations introduces two issues: (1) static boxes from accumulated scans may include objects that are either fully occluded in a given frame or located beyond the effective LiDAR/camera detection range, and (2) dynamic boxes must be associated with the correct frame using timestamps. 

To address this, we explored several filtering strategies. Our initial attempt using the timestamp range (start/end) to filter static boxes resulted in missing objects on repeated routes, while a distance-based strategy occasionally retained static boxes outside the current FoV.  Our final approach introduces per-frame LiDAR checks: static boxes are retained only if they contain points in the corresponding frame’s point cloud, thereby ensuring consistency with the instantaneous scene visibility. Dynamic boxes are preserved using frame IDs. 
\begin{figure}[htbp]
    \centering
    \begin{subfigure}{\linewidth}
        \centering
        \includegraphics[width=\linewidth]{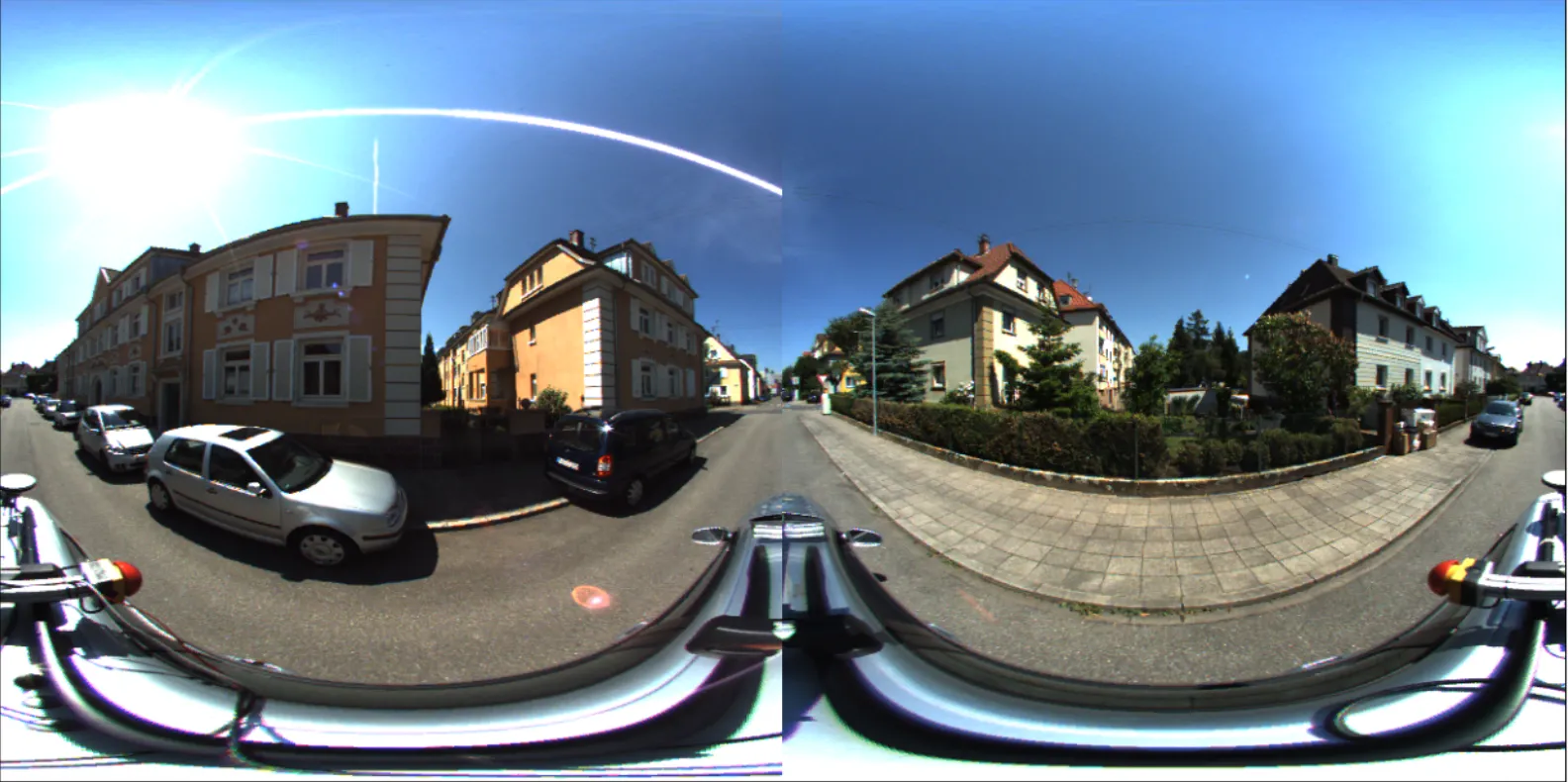}
        \caption{}
        \label{fig:kitti360_anno_equi}
    \end{subfigure}

    \begin{subfigure}[b]{0.49\linewidth}
        \centering
        \includegraphics[width=\textwidth]{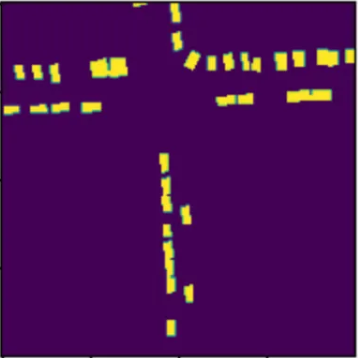}
        \caption{}
        \label{fig:kitti360_anno_bev_1}
    \end{subfigure}
    \begin{subfigure}[b]{0.49\linewidth}
        \centering
        \includegraphics[width=\textwidth]{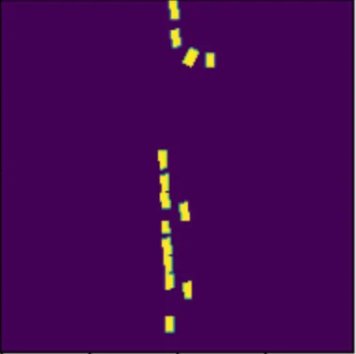}
        \caption{}
        \label{fig:kitti360_anno_bev_2}
    \end{subfigure}
    \caption{\textbf{Illustration of the annotation filtering on KITTI-360~\cite{dur360bev2025}.} (a) Equirectangular image. (b) BEV GT segmentation with distance-based strategy, where many vehicles fully occluded by buildings are still projected into the current frame. (c) BEV GT segmentation after per-frame LiDAR checks, which effectively removes such artifacts. }
    \label{fig:kitti360_anno}
\end{figure}
After this filtering, we follow the same procedure as in Dur360BEV to rasterize 3D bounding boxes into BEV ground-truth maps.

\end{document}